\documentclass[10pt,twocolumn,letterpaper]{article}
\usepackage[pagenumbers]{cvpr}
\usepackage{graphicx}
\definecolor{oursrowgray}{gray}{0.93}
\definecolor{oursrowlightgray}{gray}{0.97}
\definecolor{improvementrowblue}{RGB}{232,241,250}
\definecolor{improvementtext}{RGB}{30,74,112}
\definecolor{studentgapred}{RGB}{252,235,235}
\definecolor{studentgaptext}{RGB}{160,45,45}
\definecolor{comparisonmainblue}{RGB}{210,228,246}
\definecolor{comparisonstudentblue}{RGB}{237,245,253}
\definecolor{comparisongreen}{RGB}{230,244,232}
\definecolor{comparisongreentext}{RGB}{34,96,62}
\newsavebox{\experimenttablebox}
\newcommand{\grayrow}{}
\newcommand{\lightgrayrow}{}
\newcommand{\improvementrow}{}
\newcommand{\studentgaprow}{}
\newcommand{\comparisonrowshade}[1]{}
\makeatletter
\newcommand{\datacolorrow}[2]{%
  \begingroup
  \settowidth{\@tempdima}{#1}%
  \rlap{\kern\dimexpr\@tempdima+\tabcolsep\relax\color{#2}%
    \vrule width\dimexpr\wd\experimenttablebox-\@tempdima-\tabcolsep\relax
      height\ht\@arstrutbox depth\dp\@arstrutbox}%
  \endgroup
}
\newcommand{\experimenttable}[1]{%
  \begingroup
  \let\grayrow\relax
  \let\lightgrayrow\relax
  \let\improvementrow\relax
  \let\studentgaprow\relax
  \let\comparisonrowshade\@gobble
  \sbox{\experimenttablebox}{#1}%
  \def\grayrow{\rlap{\kern-\tabcolsep\color{oursrowgray}%
    \vrule width\wd\experimenttablebox height\ht\@arstrutbox depth\dp\@arstrutbox}}%
  \def\lightgrayrow{\rlap{\kern-\tabcolsep\color{oursrowlightgray}%
    \vrule width\wd\experimenttablebox height\ht\@arstrutbox depth\dp\@arstrutbox}}%
  \def\improvementrow{\rlap{\kern-\tabcolsep\color{improvementrowblue}%
    \vrule width\wd\experimenttablebox height\ht\@arstrutbox depth\dp\@arstrutbox}}%
  \def\studentgaprow{\rlap{\kern-\tabcolsep\color{studentgapred}%
    \vrule width\wd\experimenttablebox height\ht\@arstrutbox depth\dp\@arstrutbox}}%
  \def\comparisonrowshade##1{\rlap{\color{##1}%
    \vrule width\wd\experimenttablebox height\ht\@arstrutbox depth\dp\@arstrutbox}}%
  #1%
  \endgroup
}
\makeatother

\newcommand{\FoMNetCDFFloat}{%
\begin{figure*}[t]
\centering
\includegraphics[width=\textwidth]{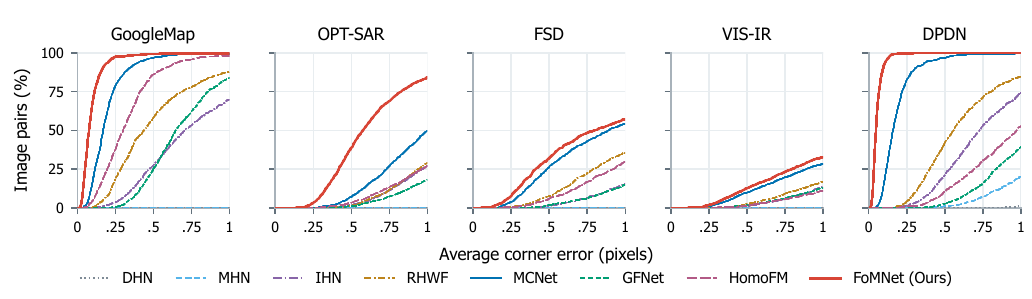}
\caption{Cumulative recall of average corner error (ACE) on five cross-modal datasets. Each curve shows the proportion of image pairs with ACE below a given threshold. Higher curves indicate more accurate homography estimation.}
\label{fig:ace_cdf}
\end{figure*}
}

\newcommand{\FoMNetQualitativeFloat}{%
\begin{figure*}[t]
\centering
\includegraphics[width=\textwidth]{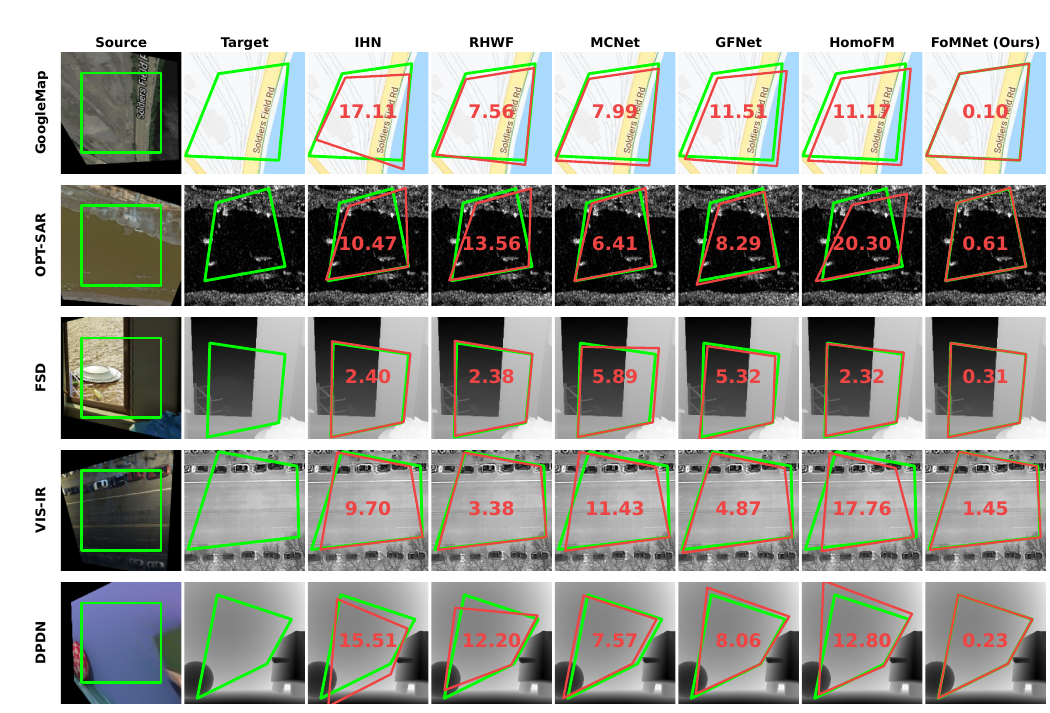}
\caption{Qualitative comparison on five modality pairs. Green and red quadrilaterals denote ground truth and predictions; red annotations report per-example ACE.  All methods share the same input pair and geometric perturbation.}
\label{fig:qualitative}
\end{figure*}
}

\newcommand{\FoMNetCorrelationFloat}{%
\begin{figure}[t]
\centering
\includegraphics[width=\columnwidth]{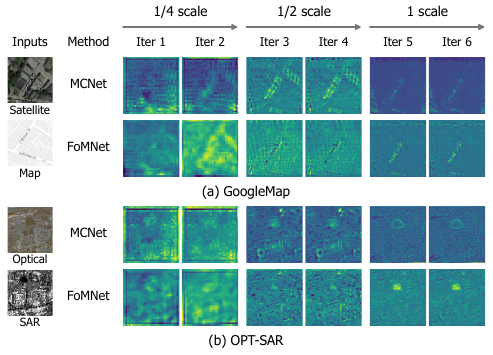}
\caption{Evolution of correlation responses in MCNet and FoMNet on GoogleMap and OPT-SAR. Columns show successive refinement iterations, with two iterations at each scale $s\in\{\tfrac14,\tfrac12,1\}$. Each map shows channel-averaged correlation responses with independent min--max normalization.}
\label{fig:correlation_evolution}
\end{figure}
}

\newcommand{\FoMNetAblationFloat}{%
\begin{table}[tbp]
\centering
\caption{Ablation studies on OPT-SAR. }
\label{tab:progressive_ablation}
\small
\setlength{\tabcolsep}{4pt}
\colorlet{oursrowgray}{comparisonmainblue}
\experimenttable{%
\begin{tabular}{llccc}
\toprule
& VFM Features & Upsampling &
\begin{tabular}[c]{@{}c@{}}
Correlation\\
Computation
\end{tabular}
& MACE $\downarrow$ \\
\midrule
1 & \hfill$\times$\hfill\null & --- & Local & 1.36 \\
2 & \hfill$\checkmark$\hfill\null & Bicubic & Local & 1.23 \\
3 & \hfill$\checkmark$\hfill\null & PFU & Local & 0.94 \\
\grayrow
Ours & \hfill$\checkmark$\hfill\null & PFU & CCR & \textbf{0.81} \\
\bottomrule
\end{tabular}%
}
\end{table}
}

\newcommand{\FoMNetDistillationFloat}{%
\begin{table}[tbp]
\centering
\caption{Effect of distillation on MACE. Colored rows show excess MACE over FoMNet (lower is better). Bicubic student uses FoMNet-D's architecture without distillation. GM, OS, and VI denote GoogleMap, OPT-SAR, and VIS-IR, respectively.}
\label{tab:distillation_accuracy}
\small
\setlength{\tabcolsep}{3pt}
\experimenttable{%
\begin{tabular*}{\columnwidth}{@{\extracolsep{\fill}}lccccc}
\toprule
Model & GM $\downarrow$ & OS $\downarrow$ & FSD $\downarrow$ & VI $\downarrow$ & DPDN $\downarrow$ \\
\midrule
FoMNet & \textbf{0.13} & \textbf{0.81} & \textbf{2.79} & \textbf{2.70} & \textbf{0.06} \\
\midrule
Bicubic student & 0.20 & 1.10 & 3.18 & 3.03 & 0.16 \\
\datacolorrow{Bicubic student}{studentgapred}
\scriptsize\textcolor{studentgaptext}{Gap to FoMNet $\downarrow$} & \scriptsize\textcolor{studentgaptext}{(+0.07)} & \scriptsize\textcolor{studentgaptext}{(+0.29)} & \scriptsize\textcolor{studentgaptext}{(+0.39)} & \scriptsize\textcolor{studentgaptext}{(+0.33)} & \scriptsize\textcolor{studentgaptext}{(+0.10)} \\
FoMNet-D & 0.15 & 1.06 & 2.81 & 2.81 & 0.08 \\
\datacolorrow{Bicubic student}{improvementrowblue}
\scriptsize\textcolor{improvementtext}{Gap to FoMNet $\downarrow$} & \scriptsize\textcolor{improvementtext}{(+0.02)} & \scriptsize\textcolor{improvementtext}{(+0.25)} & \scriptsize\textcolor{improvementtext}{(+0.02)} & \scriptsize\textcolor{improvementtext}{(+0.11)} & \scriptsize\textcolor{improvementtext}{(+0.02)} \\
\bottomrule
\end{tabular*}%
}
\end{table}
}

\newcommand{\FoMNetEfficiencyFloat}{%
\begin{table}[t]
\centering
\caption{Efficiency on pairs of $128\times128$ images. Trainable parameters and network MACs are reported in millions and billions, respectively; latency is in ms per image pair. MACs include frozen backbones but exclude geometric solvers and post-network readout. Latency is mean $\pm$ standard deviation over three timing rounds on an RTX 4090 with FP32 networks and batch size one. $\dagger$ GFNet and HomoFM are our modified versions for $128\times128$ inputs.}
\label{tab:efficiency}
\footnotesize
\setlength{\tabcolsep}{2pt}
\experimenttable{%
\begin{tabular*}{\columnwidth}{@{\extracolsep{\fill}}lccc@{}}
\toprule
Method & Trainable Parameters & MACs & Latency \\
\midrule    
MCNet & 0.85 & 4.90 & $16.50\pm0.11$ \\
RHWF & 1.34 & 14.71 & $48.65\pm1.48$ \\
GFNet$^{\dagger}$ & 3.87 & 51.78 & $51.37\pm1.87$ \\
HomoFM$^{\dagger}$ & 3.53 & 26.54 & $69.24\pm1.11$ \\
\midrule
\comparisonrowshade{comparisonmainblue}
FoMNet & 5.75 & 97.07 & $68.85\pm2.06$ \\
\comparisonrowshade{comparisonstudentblue}
FoMNet-D & 5.72 & 96.46 & $45.42\pm1.14$ \\
\bottomrule
\end{tabular*}%
}
\end{table}
}

\definecolor{cvprblue}{rgb}{0.21,0.49,0.74}
\usepackage[pagebackref,breaklinks,colorlinks,allcolors=cvprblue]{hyperref}
\def\confName{CVPR}
\def\confYear{2027}
\title{FoMNet: Adaptive Kernels Meet Contextual Matching for Cross-Modal Homography Estimation}

\author{%
Zhuoran Yang$^{1}$, Zhenyu Yi$^{2}$, Runmin Zhang$^{1}$, Linshan Li$^{1}$, Yiming Li$^{1}$\\
Xiaokai Bai$^{1}$, Xiaohan Zhang$^{1}$, Hui-liang Shen$^{1,5}$, Si-Yuan Cao$^{3,4,5}$\\
$^{1}$College of Information Science and Electronic Engineering, Zhejiang University\\
$^{2}$School of Biomedical Engineering, Shanghai Jiao Tong University\\
$^{3}$Ningbo Global Innovation Center, Zhejiang University\\
$^{4}$NingboTech University \quad $^{5}$Jinhua Institute, Zhejiang University\\
{\tt\small yangzhr2025@zju.edu.cn}
}

\begin{document}
\maketitle
\begin{abstract}
Cross-modal homography estimation requires reliable correspondences despite appearance differences between sensing modalities. Multi-scale refinement progressively reduces the search range, but its accuracy remains limited by the ability of matching features to distinguish nearby candidates. The vision foundation model (VFM) provides complementary representations, yet integrating these features through fixed interpolation leaves their spatial aggregation independent of image content and the correspondence objective. We propose the Foundation Model-Enhanced Matching Network (FoMNet), which combines learned spatial adaptation of VFM features with contextual correspondence estimation. Patch Feature Upsampling (PFU) progressively predicts location-specific kernels to aggregate neighboring features on upsampled grids, preparing VFM representations for fusion with CNN features at different matching scales. Contextual Correlation Refinement (CCR) then jointly processes source descriptors and local correlations with spatial context to refine candidate scores before each homography update. To reduce upsampling overhead, we develop FoMNet-D, which replaces PFU with bicubic interpolation and uses multi-scale feature-relation distillation to recover accuracy. Experiments on five cross-modal datasets show that FoMNet achieves the lowest mean average corner error (MACE) among the evaluated methods, reducing MACE by 10.6\%--72.7\% relative to MCNet. FoMNet-D retains competitive accuracy while reducing inference latency by 34.0\% relative to FoMNet.
\end{abstract}
\section{Introduction}\label{sec:introduction}

Homography estimation establishes a projective mapping between two views of a planar scene and supports image registration, stitching, and multi-sensor fusion. Cross-modal alignment allows complementary observations from different sensors to be combined. However, different imaging mechanisms can substantially change the intensity, texture, and local contrast of corresponding regions~\cite{kim2015dasc,li2020rift}, making reliable correspondence estimation challenging.

\begin{figure}[t]
\centering
\includegraphics[width=\linewidth]{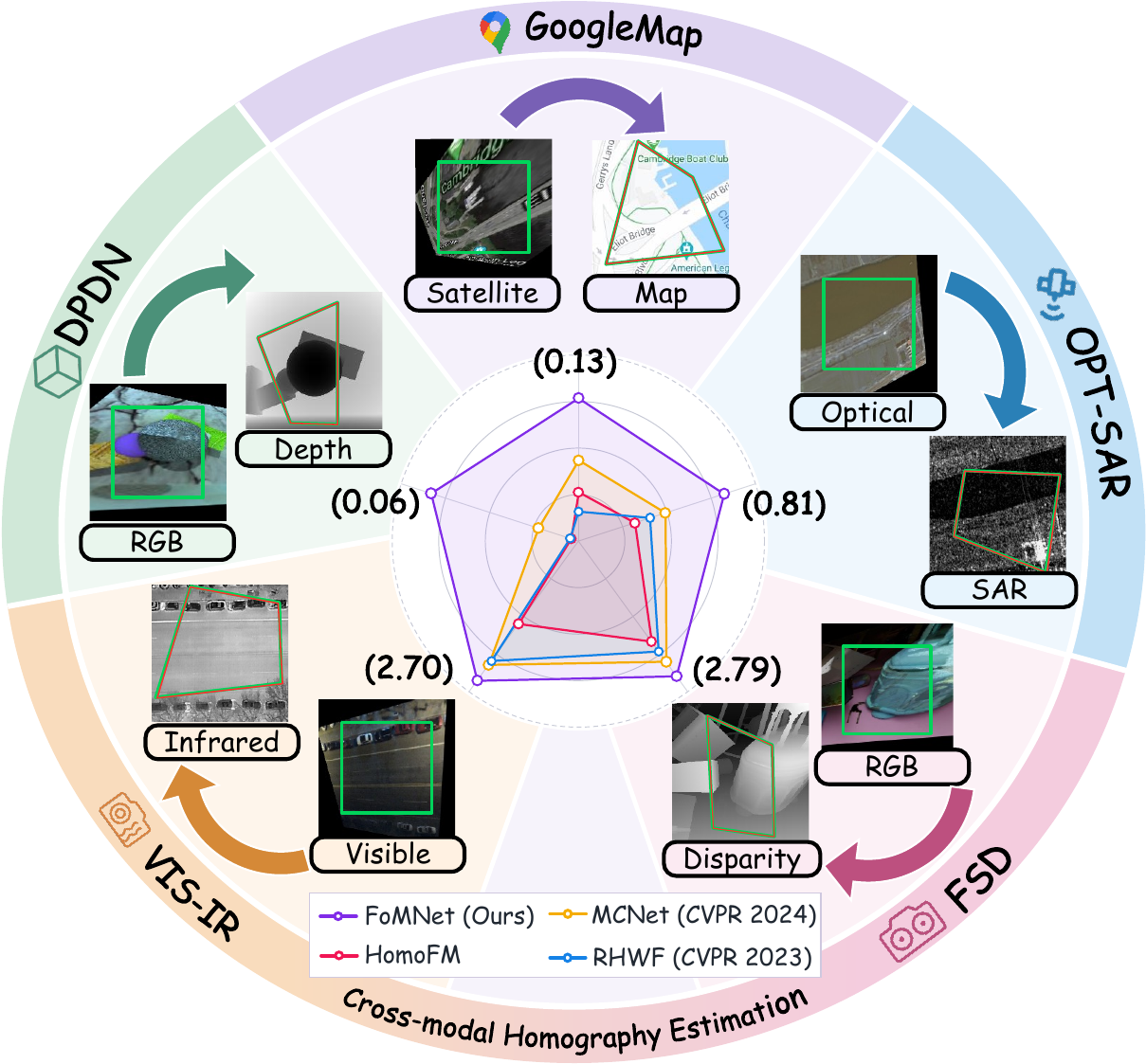}
\caption{Cross-modal homography estimation on five datasets.
Image pairs illustrate the modality differences and alignment task;
green and red quadrilaterals indicate ground truth and FoMNet
predictions, respectively. The radar plot compares MACE with
three existing methods, with larger radii indicating lower errors.}
\label{fig:teaser}
\end{figure}

Learning-based methods increasingly address homography estimation through explicit matching and iterative refinement. IHN, RHWF, and MCNet use feature correlations, cross-image interaction, and successive geometric updates to improve accuracy~\cite{cao2022ihn,cao2023rhwf,zhu2024mcnet}. Multi-scale refinement decomposes large displacements into progressively smaller corrections. Its effectiveness, however, depends on whether the features distinguish the correct correspondence within each search neighborhood. Cross-modal appearance differences can make nearby candidates difficult to separate even after the search range has been reduced. Refining the search therefore requires representations that better support correspondence estimation across modalities.

The vision foundation model (VFM) offers complementary representations learned through large-scale pretraining. RoMa combines DINOv2 features with task-specific CNN features for dense matching~\cite{oquab2023dinov2,edstedt2024roma}, while GFNet incorporates pretrained representations into homography estimation through a lightweight CNN feature pyramid~\cite{zhang2025gfnet}. Integrating these representations into multi-scale estimation requires adapting patch-level VFM features to correspondence decisions at finer matching scales.

Existing integration strategies commonly use interpolation to reconcile the spatial resolutions of VFM and CNN features. GFNet, for example, bilinearly resizes its processed VFM features before injecting them into the CNN pyramid~\cite{zhang2025gfnet}. Although subsequent feature fusion is learned, the contribution of neighboring VFM features to each resized location is determined by fixed interpolation weights. This separates spatial resizing from the correspondence objective. For cross-modal matching, where local appearance cues may be inconsistent, merely providing pretrained features at a larger spatial size may leave their potential contribution to precise localization insufficiently explored. The spatial aggregation of VFM features should therefore be considered part of the learned matching representation.

We propose the Foundation Model-Enhanced Matching Network (FoMNet), a multi-scale framework for cross-modal homography estimation (Fig.~\ref{fig:architecture}). At its core, Patch Feature Upsampling (PFU) adapts VFM features through location-specific kernel prediction and aggregation. Each stage constructs a local kernel from image-derived similarities and a spatial prior, and refines its weights through a learned residual mapping~\cite{kopf2007jbu,fu2024featup}. The kernel then aggregates neighboring features on the upsampled grid, allowing their contributions to vary with image content. Cascaded PFU stages progressively construct VFM representations at different matching resolutions. These representations are projected and fused with CNN features, with adaptation and fusion trained within the homography estimation framework.

The fused representations must also support the assessment of candidate matches. Local correlations express pairwise similarity, while source descriptors and spatial context provide additional evidence when candidate scores are ambiguous. Contextual Correlation Refinement (CCR) jointly processes these signals under the current transformation, using local interaction and pooled global context to correct correlation scores before each homography update. PFU constructs matching features through adaptive spatial aggregation, while CCR uses those features to refine correspondence decisions.

Kernel prediction and aggregation introduce additional computation. FoMNet-D replaces PFU with bicubic interpolation while retaining the VFM backbone and matching framework. Multi-scale feature-relation distillation transfers the teacher's within-image feature relations to the student, reducing the accuracy loss from simpler upsampling~\cite{hinton2015distilling,liu2019structuredkd}. Distillation is required only during training.

Experiments on GoogleMap, OPT-SAR, FSD, VIS-IR, and DPDN evaluate FoMNet across five cross-modal datasets (Fig.~\ref{fig:teaser}). FoMNet achieves the lowest mean average corner error (MACE) among the evaluated methods on all five datasets, reducing MACE relative to MCNet by 43.5\%, 40.4\%, 10.6\%, 11.2\%, and 72.7\%, respectively. FoMNet-D retains competitive accuracy while reducing inference latency by 34.0\% relative to FoMNet. Our main contributions are as follows:

\begin{itemize}
    \item We propose FoMNet, a multi-scale framework for cross-modal homography estimation that combines VFM feature adaptation with contextual correspondence estimation, achieving the lowest MACE among the evaluated methods on five cross-modal datasets.

    \item We develop PFU to progressively adapt VFM features through location-specific kernel prediction and aggregation, supporting their fusion with CNN features at different matching scales.

    \item We introduce CCR to refine candidate correlation scores by jointly processing source descriptors and local correlations with local and global spatial context.

    \item We develop multi-scale feature-relation distillation for a student with simpler feature upsampling, retaining competitive accuracy while reducing inference latency by 34.0\%.
\end{itemize}

\section{Related Work}\label{sec:related}

\subsection{Deep Homography Estimation}

Deep homography estimation~\cite{nguyen2018udh,zhang2020content,ye2021baseshomo,hong2022homogan} has evolved from direct transformation regression to explicit matching and iterative refinement. Early approaches predict corner offsets from paired images, while subsequent methods improve accuracy through cascaded prediction and learned iterative updates \cite{detone2016,le2020mhn,cao2022ihn}. Feature correlations~\cite{teed2020raft,cho2021cats}, attention-based interaction~\cite{sun2021loftr,wang2024efficientloftr,liu2023geoformer}, and multi-scale search enable increasingly precise geometric estimation \cite{shao2021localtrans,cao2023rhwf,zhu2024mcnet}. However, the effectiveness of these refinement mechanisms depends on the reliability of the underlying correspondences. Progressively narrowing the spatial search does not by itself resolve feature discrepancies caused by different imaging modalities.

Cross-modal methods address these discrepancies through representation learning and alignment objectives~\cite{you2026ccnet,ren2025minima}. Learned feature spaces and optimization constraints improve the suitability of Lucas--Kanade alignment for multimodal inputs \cite{zhao2021dlkfm,zhang2023prise}. Intra-modal self-supervision and decomposed optimization further provide training strategies that reduce reliance on direct cross-modal photometric consistency \cite{zhang2024scpnet,song2024alto,yu2025sshnet}. These advances improve how alignment is learned, but reliable correspondence estimation remains challenging when modality differences obscure local appearance cues. This limitation motivates representations that complement task-specific features with transferable information from large-scale visual pretraining.

\subsection{Foundation Features for Geometric Matching}

Foundation features have increasingly been adopted in geometric matching to complement representations learned from task-specific data \cite{edstedt2024roma,liu2024srmatcher,zhang2025gfnet,he2026homofm}. Their integration commonly combines coarse pretrained representations with finer CNN features to balance robustness to appearance variations and localization accuracy \cite{edstedt2024roma,zhang2025gfnet,chen2023vitadapter}. This strategy has extended from dense correspondence estimation to homography estimation, demonstrating the value of pretrained representations for geometric alignment.

However, constructing a feature hierarchy does not automatically make coarse foundation representations suitable for matching at every scale. In GFNet and HomoFM, interpolation aligns the spatial resolution of foundation features before fusion~\cite{zhang2025gfnet,he2026homofm}, but its fixed aggregation weights cannot adapt to image structure or the matching objective~\cite{fu2024featup,ranzinger2025featsharp,couairon2025jafar,wimmer2026anyup}. These limitations highlight the need to adapt foundation features to the spatial precision and correspondence discrimination required by multi-scale cross-modal matching.

\section{Method}\label{sec:method}

\begin{figure*}[t]
\centering
\includegraphics[width=1.0\textwidth]{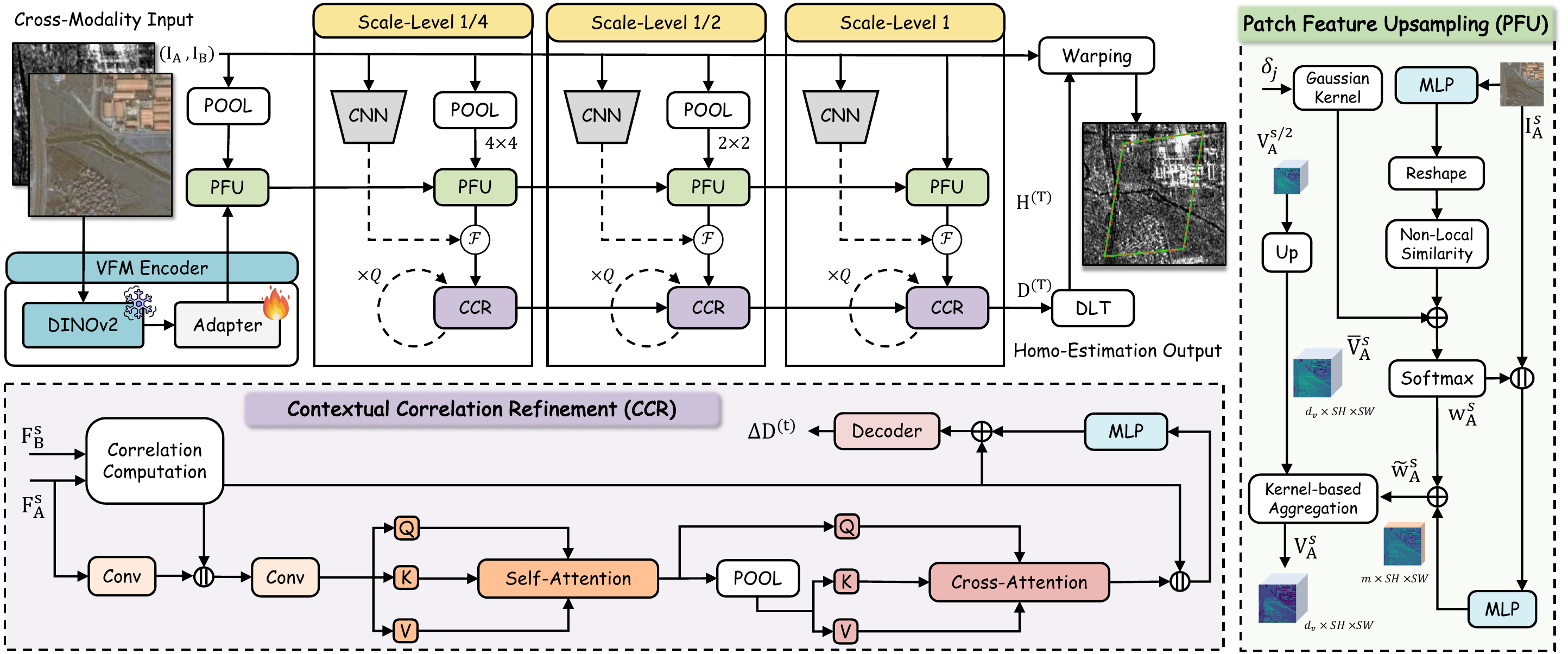}
\caption{
Overview of FoMNet.
A shared encoder progressively adapts frozen DINOv2 features through PFU, with POOL providing input images at the required scales.
At each matching scale, $\mathcal F$ projects the adapted VFM features and fuses them with CNN features.
CCR then jointly processes source descriptors and local correlations using local interactions and global context constructed by spatial pooling (POOL).
The refined correlations support corner-displacement updates over $Q$ iterations per scale.
After the three matching scales, DLT (Direct Linear Transform) recovers the final homography from the reference corners and their accumulated displacements, and Warping applies this transformation to the input image.
}
\label{fig:architecture}
\end{figure*}

\subsection{Overview}

Given a source image \(\mathbf I_{\mathrm A}\) and a target image \(\mathbf I_{\mathrm B}\), FoMNet estimates their homography \(\mathbf H\) through multi-scale feature encoding and contextual correlation refinement (Fig.~\ref{fig:architecture}). A shared encoder progressively adapts frozen foundation features and fuses them with CNN features at three matching scales. The fused features are reused across coarse-to-fine iterations, where CCR refines local correlations using source descriptors and spatial context before each homography update.

\subsection{Multi-Scale Feature Adaptation and Fusion}

Our encoder adapts frozen foundation features and fuses them with CNN features at three matching scales. We describe the process for modality A; the same encoder is applied to modality B.

\paragraph{Pretrained feature extraction.} Given an input patch \(\mathbf I_{\mathrm A}\) of size \(H\times W\), we extract patch features \(\mathbf V_{\mathrm A}^{\mathrm{pre}}\) using a frozen DINOv2 backbone \cite{oquab2023dinov2}. A trainable adapter maps \(\mathbf V_{\mathrm A}^{\mathrm{pre}}\) to \(\mathbf V_{\mathrm A}^{1/16}\) at \(H/16\times W/16\) resolution for subsequent progressive upsampling.

\paragraph{Progressive feature upsampling.} Starting from \(\mathbf V_{\mathrm A}^{1/16}\), we progressively adapt the coarse foundation features to the required matching resolutions through four cascaded patch feature upsampling (PFU) modules. We describe a single module from relative scale \(s/2\) to \(s\), where \(s\in\{\tfrac18,\tfrac14,\tfrac12,1\}\).

Each PFU takes the preceding features \(\mathbf V_{\mathrm A}^{s/2}\) and a guidance image \(\mathbf I_{\mathrm A}^{s}\) of size \(sH\times sW\) as input. The guidance images are obtained by average pooling the input image to the corresponding resolutions, with the original image used directly at the finest scale, i.e., \(\mathbf I_{\mathrm A}^{1}=\mathbf I_{\mathrm A}\). Within PFU, we upsample \(\mathbf V_{\mathrm A}^{s/2}\) by a factor of two and extract guidance embeddings through a pointwise network \(\phi^{s}\):

\begin{equation}
\overline{\mathbf V}_{\mathrm A}^{s}=\operatorname{Up}_{2}(\mathbf V_{\mathrm A}^{s/2}),\qquad
\mathbf E_{\mathrm A}^{s}=\phi^{s}(\mathbf I_{\mathrm A}^{s}).
\label{eq:method2}
\end{equation}

Here, \(\overline{\mathbf V}_{\mathrm A}^{s}(\mathbf p)\in\mathbb R^{d_v}\) and \(\mathbf E_{\mathrm A}^{s}(\mathbf p)\in\mathbb R^{d_e}\), where \(d_v\) and \(d_e\) denote the adapted feature and guidance embedding dimensions, respectively.

At each position \(\mathbf p\), we compute guidance similarities through softmax, multiply them by a Gaussian spatial kernel, and normalize the resulting aggregation weights~\cite{kopf2007jbu,fu2024featup}:

\begin{equation}
w_{\mathrm A,j}^{s}(\mathbf p)
=\frac{g^{s}(\boldsymbol\delta_j)}{Z_{\mathrm A}^{s}(\mathbf p)}
\operatorname{softmax}\!\left(
\mathbf E_{\mathrm A}^{s}(\mathbf p)^\top
\mathbf E_{\mathrm A}^{s}(\mathbf p+\boldsymbol\delta_j)
\right).
\label{eq:method3}
\end{equation}

Here, \(\Omega=\{-r,\ldots,r\}^{2}\) denotes the neighborhood offsets with radius \(r\) and \(m=|\Omega|\). The offsets \(\boldsymbol\delta_1,\ldots,\boldsymbol\delta_m\) follow a fixed ordering, and softmax is applied over the \(m\) neighbors. The Gaussian spatial kernel \(g^{s}>0\) has a learned bandwidth, and \(Z_{\mathrm A}^{s}(\mathbf p)\) normalizes the weights to sum to one. We refine the resulting weight vector \(\mathbf w_{\mathrm A}^{s}(\mathbf p)\in\mathbb R^{m}\) through a guidance-conditioned residual mapping:

\begin{equation}
\widetilde{\mathbf w}_{\mathrm A}^{s}(\mathbf p)
=\mathbf w_{\mathrm A}^{s}(\mathbf p)
+\alpha\,\psi^{s}\!\left(\mathbf w_{\mathrm A}^{s}(\mathbf p)\mathbin\Vert \mathbf I_{\mathrm A}^{s}(\mathbf p)\right),
\label{eq:method5}
\end{equation}

where \(\Vert\) denotes channel concatenation, \(\psi^{s}:\mathbb R^{m+3}\to\mathbb R^{m}\) is a pointwise kernel prediction network, and \(\alpha\) is a fixed residual scaling factor. The refined kernel \(\widetilde{\mathbf w}_{\mathrm A}^{s}(\mathbf p)\in\mathbb R^{m}\) provides scalar weights for kernel-based aggregation of the \(d_v\)-dimensional neighboring feature vectors:

\begin{equation}
\mathbf V_{\mathrm A}^{s}(\mathbf p)
=\sum_{j=1}^{m}
\widetilde w_{\mathrm A,j}^{s}(\mathbf p)\,
\overline{\mathbf V}_{\mathrm A}^{s}(\mathbf p+\boldsymbol\delta_j).
\label{eq:method6}
\end{equation}

The output \(\mathbf V_{\mathrm A}^{s}(\mathbf p)\in\mathbb R^{d_v}\) retains the feature dimension. The aggregation kernel is shared across channels and applied without additional normalization. The last three PFU stages supply features at \(H/4\times W/4\), \(H/2\times W/2\), and \(H\times W\) resolution.

\paragraph{Feature fusion.} At each matching scale \(s\in\mathcal S=\{\tfrac14,\tfrac12,1\}\), a fusion module integrates the upsampled foundation features \(\mathbf V_{\mathrm A}^{s}\) with the CNN features \(\mathbf C_{\mathrm A}^{s}\) at the same resolution. The module first projects the foundation features to the CNN feature dimension through a scale-specific matching head, then concatenates and fuses the two representations through convolutional layers. At the finest scale, a residual channel refinement precedes projection. The fused matching features are

\begin{equation}
\mathbf F_{\mathrm A}^{s}
=\mathcal F^{s}\!\left(\mathbf C_{\mathrm A}^{s},\mathbf V_{\mathrm A}^{s}\right).
\label{eq:method7}
\end{equation}

Here, \(\mathcal F^{s}\) denotes the feature projection and fusion module at scale \(s\). The fused features \(\mathbf F_{\mathrm A}^{s}\) support both inter-image correlation computation and source descriptor encoding in CCR. Modality B follows the same process with shared encoder parameters, yielding \(\mathbf F_{\mathrm B}^{s}\).

\begin{table*}[t]
\centering
\caption{Homography estimation results on five cross-modal datasets. MACE is measured in pixels (lower is better), and AUC@$t$ is reported in percent (higher is better). FoMNet-D is the distilled variant. Bold and underlined values indicate the best and second-best results, respectively, with FoMNet-D excluded from the second-best ranking. Improvement (\%) denotes FoMNet's relative MACE reduction and AUC gains over the underlined second-best results.}
\label{tab:comparison}
\scriptsize
\setlength{\tabcolsep}{0.8pt}
\renewcommand{\arraystretch}{1.12}
\newcommand{\improvementvalue}[1]{{\tiny\textcolor{comparisongreentext}{+#1}}}
\experimenttable{%
\begin{tabular*}{\textwidth}{@{\extracolsep{\fill}}lrrrrrrrrrrrrrrrrrrrr@{}}
\toprule
Method & \multicolumn{4}{c}{GoogleMap} & \multicolumn{4}{c}{OPT-SAR} & \multicolumn{4}{c}{FSD} & \multicolumn{4}{c}{VIS-IR} & \multicolumn{4}{c}{DPDN} \\
\cmidrule(lr){2-5}\cmidrule(lr){6-9}\cmidrule(lr){10-13}\cmidrule(lr){14-17}\cmidrule(lr){18-21}
 & MACE\,$\scriptstyle\downarrow$ & AUC$_1$ & AUC$_3$ & AUC$_5$ & MACE\,$\scriptstyle\downarrow$ & AUC$_1$ & AUC$_3$ & AUC$_5$ & MACE\,$\scriptstyle\downarrow$ & AUC$_1$ & AUC$_3$ & AUC$_5$ & MACE\,$\scriptstyle\downarrow$ & AUC$_1$ & AUC$_3$ & AUC$_5$ & MACE\,$\scriptstyle\downarrow$ & AUC$_1$ & AUC$_3$ & AUC$_5$ \\
\midrule
DHN \cite{detone2016} & 5.15 & 0.00 & 3.43 & 17.86 & 9.14 & 0.00 & 0.18 & 2.04 & 12.4 & 0.00 & 0.05 & 0.69 & 8.93 & 0.00 & 0.23 & 3.09 & 4.92 & 0.14 & 17.19 & 41.44 \\
MHN \cite{le2020mhn} & 3.55 & 0.08 & 11.12 & 34.54 & 5.69 & 0.00 & 1.33 & 11.44 & 7.91 & 0.00 & 0.81 & 5.97 & 6.46 & 0.02 & 2.74 & 14.44 & 2.95 & 4.29 & 46.68 & 66.68 \\
IHN \cite{cao2022ihn} & 0.98 & 29.72 & 69.46 & 80.93 & 1.68 & 6.37 & 50.01 & 68.28 & 3.91 & 3.75 & 33.66 & 50.06 & 3.63 & 3.32 & 29.80 & 48.17 & 1.17 & 27.56 & 72.01 & 83.00 \\
RHWF \cite{cao2023rhwf} & 0.63 & 49.48 & 80.84 & 88.18 & 1.65 & 6.65 & 51.43 & 69.36 & 3.41 & 11.97 & 44.56 & 58.65 & 3.14 & 4.80 & 33.14 & 51.34 & 1.06 & 39.85 & 77.66 & 86.35 \\
MCNet \cite{zhu2024mcnet} & \underline{0.23} & \underline{80.13} & \underline{93.08} & \underline{95.67} & \underline{1.36} & \underline{14.31} & \underline{59.72} & \underline{74.52} & \underline{3.12} & \underline{24.93} & \underline{54.77} & \underline{65.50} & \underline{3.04} & \underline{11.09} & \underline{40.14} & \underline{56.69} & \underline{0.22} & \underline{81.08} & \underline{93.55} & \underline{96.05} \\
GFNet \cite{zhang2025gfnet} & 0.84 & 32.38 & 74.70 & 84.49 & 2.52 & 4.18 & 38.71 & 57.61 & 4.15 & 3.73 & 30.02 & 46.43 & 4.11 & 3.50 & 28.22 & 45.86 & 1.38 & 9.31 & 56.24 & 73.03 \\
HomoFM \cite{he2026homofm} & 0.38 & 66.81 & 88.43 & 92.87 & 2.10 & 7.33 & 45.93 & 63.74 & 3.75 & 9.12 & 41.23 & 55.12 & 4.55 & 3.09 & 26.51 & 44.22 & 1.18 & 16.28 & 62.41 & 76.68 \\
\midrule
\comparisonrowshade{comparisonmainblue}
FoMNet (Ours) & \textbf{0.13} & \textbf{90.40} & \textbf{96.57} & \textbf{97.81} & \textbf{0.81} & \textbf{38.19} & \textbf{75.77} & \textbf{84.79} & \textbf{2.79} & 29.16 & \textbf{57.41} & \textbf{67.51} & \textbf{2.70} & 13.36 & 43.84 & \textbf{60.13} & \textbf{0.06} & \textbf{94.41} & \textbf{98.07} & \textbf{98.80} \\
\comparisonrowshade{comparisongreen}\textcolor{comparisongreentext}{\raisebox{-0.8pt}{\includegraphics[height=6.5pt]{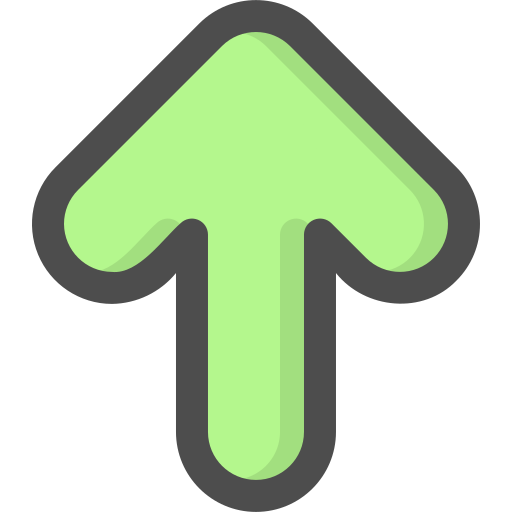}}\,\textbf{Improvement (\%)}} & \improvementvalue{43.48} & \improvementvalue{12.82} & \improvementvalue{3.75} & \improvementvalue{2.24} & \improvementvalue{40.44} & \improvementvalue{166.88} & \improvementvalue{26.88} & \improvementvalue{13.78} & \improvementvalue{10.58} & \improvementvalue{16.97} & \improvementvalue{4.82} & \improvementvalue{3.07} & \improvementvalue{11.18} & \improvementvalue{20.47} & \improvementvalue{9.22} & \improvementvalue{6.07} & \improvementvalue{72.73} & \improvementvalue{16.44} & \improvementvalue{4.83} & \improvementvalue{2.86} \\
\midrule
\comparisonrowshade{comparisonstudentblue}
FoMNet-D & 0.15 & 89.23 & 96.11 & 97.54 & 1.06 & 23.78 & 68.11 & 80.31 & 2.81 & \textbf{29.31} & 56.49 & 66.63 & 2.81 & \textbf{13.85} & \textbf{43.98} & 59.95 & 0.08 & 93.78 & 97.93 & 98.76 \\
\bottomrule
\end{tabular*}%
}
\end{table*}

\subsection{Contextual Correlation Refinement}

To incorporate spatial context into local matching, CCR jointly processes correlation scores and source descriptors before each homography update.

\paragraph{Local correlation construction.} We perform \(Q\) updates at each matching scale, giving \(T=3Q\) updates in total. Let \(t\in\{1,\ldots,T\}\) index these updates. At scale \(s\), we warp the target features using the current homography \(\mathbf H^{(t-1)}\), obtaining \(\widehat{\mathbf F}_{\mathrm B}^{s,(t)}\). For each source position \(\mathbf p\), the target descriptors at \(\mathbf p+\boldsymbol\delta\), \(\boldsymbol\delta\in\Lambda\), form the columns of a neighborhood matrix \(\mathcal N_{\mathbf p}^{\Lambda}(\widehat{\mathbf F}_{\mathrm B}^{s,(t)})\). The local correlation vector is computed through channel-averaged inner products:

\begin{equation}
\mathbf r^{s,(t)}(\mathbf p)
=\frac{1}{C}\mathcal N_{\mathbf p}^{\Lambda}(\widehat{\mathbf F}_{\mathrm B}^{s,(t)})^\top
\mathbf F_{\mathrm A}^{s}(\mathbf p),
\label{eq:method9}
\end{equation}

where \(C\) is the feature dimension and \(\Lambda=\{-4,\ldots,4\}^2\). Arranging these vectors over the source grid yields the correlation map \(\mathbf R^{s,(t)}\).

We jointly encode the local correlation vector and the source descriptor
\(\mathbf a_{\mathrm A}^{s}(\mathbf p)=\eta^{s}(\mathbf F_{\mathrm A}^{s}(\mathbf p))\).
Their concatenation is projected into a shared embedding space and augmented with positional encoding:

\begin{equation}
\mathbf T^{s,(t)}(\mathbf p)
=\operatorname{MLP}^{s}\!\left(\mathbf r^{s,(t)}(\mathbf p)\mathbin\Vert\mathbf a_{\mathrm A}^{s}(\mathbf p)\right)
+\mathbf P^{s}(\mathbf p),
\label{eq:method10}
\end{equation}

where \(\Vert\) denotes channel concatenation and \(\operatorname{MLP}^{s}\) denotes the token embedding projection, implemented as a single affine layer. The fixed two-dimensional sinusoidal positional encoding \(\mathbf P^{s}(\mathbf p)\) has the same dimension as the projected token and encodes source position \(\mathbf p\). Each token represents a source position, enabling subsequent spatial attention to jointly consider matching evidence and source descriptors.

\FoMNetCDFFloat

\paragraph{Local and global contextual aggregation.} We combine window-based interaction \cite{liu2021swin} with attention to spatially reduced context \cite{chu2021twins}. We first apply self-attention to \(\mathbf T^{s,(t)}\) within non-overlapping \(M\times M\) windows, yielding locally contextualized tokens \(\mathbf T_{\mathrm{loc}}^{s,(t)}\). Here, \(M\) denotes the window side length. We then spatially pool \(\mathbf T_{\mathrm{loc}}^{s,(t)}\) into an \(8\times8\) memory grid. Cross-attention uses all source positions as queries and the pooled tokens as keys and values, enabling information exchange across windows. Both attention blocks include standard normalization, feed-forward layers, and residual connections. The resulting contextual representation, denoted by \(\mathbf Z^{s,(t)}\), combines local interactions with global context while limiting the number of keys used for global attention.

\paragraph{Correlation refinement and homography update.} Given the contextual representation \(\mathbf Z^{s,(t)}(\mathbf p)\), a residual readout refines the local correlation vector:

\begin{equation}
\begin{aligned}
\widetilde{\mathbf r}^{s,(t)}(\mathbf p)
&=\mathbf r^{s,(t)}(\mathbf p)\\
&\quad+\mathcal G^{s}\!\left(\mathbf r^{s,(t)}(\mathbf p)\mathbin\Vert \mathbf Z^{s,(t)}(\mathbf p)\right).
\end{aligned}
\label{eq:method14}
\end{equation}

The final readout layer is initialized to zero, preserving the original correlations at initialization. Arranging the refined vectors over the source grid yields \(\widetilde{\mathbf R}^{s,(t)}\).

Let \(\mathbf D^{(t)}\in\mathbb R^{4\times2}\) denote the accumulated displacements of the four reference corners after \(t\) updates. At each iteration, the corner decoder takes \(\widetilde{\mathbf R}^{s,(t)}\) as input and predicts residual corner displacements \(\Delta\mathbf D^{(t)}\), accumulated as \(\mathbf D^{(t)}=\mathbf D^{(t-1)}+\Delta\mathbf D^{(t)}\), with \(\mathbf D^{(0)}=\mathbf 0\). The updated corner displacements define the current homography \(\mathbf H^{(t)}\), which guides target feature alignment in the next iteration. We recover the final homography \(\mathbf H\) from the corner correspondences defined by \(\mathbf D^{(T)}\).

\FoMNetQualitativeFloat

\subsection{Distillation for Simplified Feature Upsampling}\label{sec:distillation}

To reduce the computational overhead of feature upsampling, FoMNet-D replaces the PFU modules with bicubic interpolation while retaining the frozen DINOv2 backbone and the matching pipeline. The trained FoMNet serves as a frozen teacher, whose spatial similarity structure guides the student's fused matching representations at three scales.

Replacing PFU with bicubic interpolation simplifies feature upsampling, but retaining the teacher's spatial feature structure remains a challenge. Prior studies on spatial affinity supervision provide a useful perspective by transferring relationships between feature locations \cite{liu2019structuredkd,huang2025loftup}. Motivated by this perspective, we supervise the student using within-image pairwise feature similarities at each matching scale.
Let \(\mathbf F_{q,i}^{s}\) denote the fused features of model \(q\in\{\mathrm T,\mathrm S\}\), identifying the teacher or student, for image \(i\in\{\mathrm A,\mathrm B\}\). For each image \(i\) and scale \(s\), we sample \(K\) positions shared by the teacher and student. Let \(\overline{\mathbf f}_{q,i,u}^{s}\) denote the \(\ell_2\)-normalized feature at sampled position \(u\in\{1,\ldots,K\}\). Their pairwise similarities are

\begin{equation}
\mathbf A_{q,i}^{s}(u,v)
=(\overline{\mathbf f}_{q,i,u}^{s})^\top
\overline{\mathbf f}_{q,i,v}^{s}.
\label{eq:distillation_relations}
\end{equation}

The distillation loss aligns the off-diagonal relations:

\begin{equation}
\begin{aligned}
\mathcal L_{\mathrm{KD}}
&=\sum_{s\in\mathcal S}\frac{1}{2K(K-1)}
\sum_{i\in\{\mathrm A,\mathrm B\}}\sum_{u\ne v}\\
&\quad\left[
\mathbf A_{\mathrm S,i}^{s}(u,v)
-\mathbf A_{\mathrm T,i}^{s}(u,v)
\right]^2.
\end{aligned}
\label{eq:method16}
\end{equation}

This objective transfers within-image similarity structure, while cross-image alignment remains supervised by the homography loss. Only the student is used at inference.

\subsection{Training Objectives}

We supervise all \(T\) corner updates using a weighted sequence loss:

\begin{equation}
\mathcal L_{\mathrm{homo}}
=\sum_{t=1}^{T}\gamma^{T-t}
\frac{1}{8}\|\mathbf D^{(t)}-\mathbf D_{\mathrm{GT}}\|_1,
\label{eq:method17}
\end{equation}

where \(\mathbf D_{\mathrm{GT}}\) denotes the ground-truth corner displacements and the \(\ell_1\) term sums absolute errors over the eight corner coordinates. Pixel reconstruction is an established supervision signal for visual representation learning~\cite{pathak2016context,he2022mae}. We use it as an auxiliary objective: a training-only reconstruction head predicts each input image \(\mathbf I_i\) from \(\mathbf V_i^{1/4}\), yielding an MSE loss \(\mathcal L_{\mathrm{rec}}\). The overall objective is

\begin{equation}
\mathcal L
=\mathcal L_{\mathrm{homo}}
+\lambda_{\mathrm{rec}}\mathcal L_{\mathrm{rec}}
+\lambda_{\mathrm{KD}}\mathcal L_{\mathrm{KD}},
\label{eq:method19}
\end{equation}

where \(\lambda_{\mathrm{KD}}\) is set to zero when training the full model and enabled during student distillation.

\section{Experiments}\label{sec:experiments}
\setlength{\textfloatsep}{12pt plus 2pt minus 2pt}
\setlength{\floatsep}{8pt plus 2pt minus 2pt}
\FoMNetAblationFloat

\subsection{Datasets and Experimental Setup}\label{sec:setup}

\paragraph{Datasets.} 
To comprehensively evaluate FoMNet across different modality combinations, experiments cover five cross-modal datasets: GoogleMap~\cite{zhao2021dlkfm}, OPT-SAR~\cite{li2022mcanet}, FSD~\cite{wen2025foundationstereo}, VIS-IR~\cite{sun2022drone}, and DPDN~\cite{riegler2016dpdn}, representing satellite/map, optical/SAR, RGB/disparity, visible/infrared, and RGB/depth pairs, respectively.
We follow a widely adopted data-generation protocol in deep homography estimation~\cite{detone2016,cao2022ihn,cao2023rhwf,zhu2024mcnet,zhang2024scpnet,yu2025sshnet}, generating homography-deformed image pairs from \(128\times128\) patches with random corner perturbations within \([-32,32]\) pixels.

\paragraph{Evaluation metrics.} We report mean average corner error (MACE), obtained by averaging per-pair average corner error (ACE), in the $128\times128$ evaluation coordinates. For threshold-based evaluation, AUC@$t$ is the ACE cumulative recall integrated up to $t$ by the trapezoidal rule and normalized by $t$. Lower MACE and higher AUC indicate better accuracy. Table~\ref{tab:comparison} reports MACE and AUC at 1, 3, and 5 pixels.

\paragraph{Implementation details.}
FoMNet is trained for 120,000 iterations with a batch size of 16. All methods share the same dataset splits, \(128\times128\) input resolution, homography perturbation settings, and evaluation metrics to ensure a fair comparison. GFNet~\cite{zhang2025gfnet} and HomoFM~\cite{he2026homofm} are retrained under this common experimental protocol. Additional experimental settings are provided in Appendix~A of the supplementary material.

\subsection{Comparison with Existing Methods}\label{sec:comparison}
\FoMNetDistillationFloat
\FoMNetCorrelationFloat


FoMNet achieves the lowest MACE on all five cross-modal datasets (Table~\ref{tab:comparison}). Compared with MCNet, a prior state-of-the-art method for homography estimation, FoMNet reduces MACE by 43.48\%, 40.44\%, 10.58\%, 11.18\%, and 72.73\% on GoogleMap, OPT-SAR, FSD, VIS-IR, and DPDN, respectively. The corresponding relative gains in AUC@1 are 12.82\%, 166.88\%, 16.97\%, 20.47\%, and 16.44\%. Improvements also hold at larger thresholds, with relative gains of 3.75--26.88\% in AUC@3 and 2.24--13.78\% in AUC@5. These gains span all five modality pairs and all reported metrics. FoMNet-D retains broadly comparable accuracy to its teacher while outperforming the comparison methods on all five datasets.

Figure~\ref{fig:ace_cdf} shows that FoMNet achieves higher cumulative recall than all comparison methods over most of the 0--1 pixel interval across all five datasets, indicating more accurate homography estimation under strict error thresholds.

FoMNet estimates corner locations more accurately despite the appearance differences between modalities (Fig.~\ref{fig:qualitative}). Its predicted quadrilaterals closely follow the ground truth in the illustrated examples, while the comparison methods show larger deviations along the edges and at the corners. The lower ACE values are consistent with these visual differences across all five modality pairs. Additional qualitative results are provided in Appendix~F of the supplementary material.

The two methods further differ in how their decoder-input correlation responses evolve across scales and iterations. 
Figure~\ref{fig:correlation_evolution} examines how the correlation responses supplied to the decoder evolve across scales and iterations. At $s=\tfrac14$, MCNet exhibits prominent boundary responses in both examples. FoMNet shows a more noticeable redistribution of responses within the image interior between the first two updates, particularly in GoogleMap. At $s=\tfrac12$, both methods reveal finer spatial patterns, although these remain distributed across multiple regions. At $s=1$, the elongated diagonal structure in GoogleMap and the compact central region in OPT-SAR become more distinct within FoMNet's response maps. These patterns remain similar over the final two updates, whereas MCNet exhibits less distinct interior structure in GoogleMap and more pervasive repetitive patterns in OPT-SAR.

These differences concern the spatial organization of the decoder input, rather than simply its resolution. They are consistent with FoMNet's design, which combines adapted VFM features with contextual correlation refinement throughout multi-scale estimation. For the two illustrated pairs, the final ACE decreases from 0.266 to 0.149 pixels on GoogleMap and from 0.546 to 0.295 pixels on OPT-SAR compared with MCNet. The visualization therefore complements the quantitative results by showing how changes in spatial correlation patterns accompany more accurate homography estimates.

\vspace{-0.5\baselineskip}
\subsection{Ablation Studies}\label{sec:ablations}
\FoMNetEfficiencyFloat

We conduct  ablations studies on OPT-SAR to evaluate VFM feature integration, feature upsampling, and correlation computation (Table~\ref{tab:progressive_ablation}).

\vspace{-0.5\baselineskip}
\paragraph{VFM features.}
Introducing the frozen VFM branch with bicubic upsampling reduces MACE from 1.36 to 1.23 pixels, a relative reduction of 9.56\%. This result supports integrating pretrained visual representations with CNN features to improve cross-modal homography estimation, even with simple interpolation-based upsampling.

\vspace{-0.5\baselineskip}
\paragraph{Feature upsampling.}
Replacing bicubic upsampling with PFU reduces MACE from 1.23 to 0.94 pixels, a relative reduction of 23.58\%. Although both configurations provide VFM features at the required matching scales, PFU achieves better accuracy through learned spatial adaptation. This result supports adapting VFM features for fusion and matching beyond simply aligning their spatial resolutions.

\vspace{-0.5\baselineskip}
\paragraph{Correlation computation.}
Introducing CCR reduces MACE further from 0.94 to 0.81 pixels, a relative reduction of 13.83\%. With the feature encoder unchanged, this improvement supports jointly processing local correlations and source descriptors with spatial context. CCR thus provides an additional gain beyond feature adaptation, demonstrating the complementary benefits of PFU and CCR. Ablations of CCR's internal components are provided in Appendix~B of the supplementary material.

\vspace{-0.5\baselineskip}
\subsection{Encoder Distillation and Efficiency}\label{sec:efficiency}

Distillation reduces the student's MACE on all five datasets, narrowing the accuracy gap introduced by replacing PFU with bicubic upsampling (Table~\ref{tab:distillation_accuracy}). FoMNet-D comes within 0.02 pixels of its teacher on GoogleMap, FSD, and DPDN, although a larger gap of 0.25 pixels remains on OPT-SAR. These results support distillation as an effective way to recover accuracy while retaining the simpler upsampling architecture.

Table~\ref{tab:efficiency} shows that FoMNet-D reduces inference latency by 34.0\%, from 68.85 to 45.42 ms per image pair. Although FoMNet-D has nearly the same trainable parameter count and MACs as FoMNet, it achieves substantially lower latency under the same evaluation setting. This contrast highlights the practical benefit of simplifying feature upsampling and indicates that aggregate complexity metrics alone do not fully characterize inference efficiency. Further experiments on the distillation design are provided in Appendix~D of the supplementary material.

\vspace{-0.5\baselineskip}
\section{Conclusion}\label{sec:conclusion}

We presented FoMNet for cross-modal homography estimation, adapting VFM features to multi-scale feature construction and correlation refinement. PFU progressively adapts VFM features for fusion with CNN features, while CCR uses source descriptors and spatial context to refine local correlations before each homography update. Experiments on five cross-modal datasets show that FoMNet achieves the lowest MACE among the evaluated methods, with ablations supporting the complementary benefits of PFU and CCR. Multi-scale feature-relation distillation transfers the teacher's spatial feature relations to FoMNet-D, retaining competitive accuracy with bicubic upsampling and reducing inference latency by 34.0\%. Despite this reduction, inference efficiency remains a limitation. Future work will focus on further improving inference speed while preserving estimation accuracy.

{\small
\bibliographystyle{ieeenat_fullname}
\bibliography{references}
}
\end{document}